# Autonomous Vehicle Navigation with LIDAR using Path Planning

Rahul M K, Sumukh B, Praveen L Uppunda, Vinayaka Raju, C Gururaj
B M S College of Engineering, Bengaluru, India

**Abstract:** In this paper, a complete framework for Autonomous Self Driving is implemented. LIDAR, Camera and IMU (Inertial Measurement Unit) sensors are used together. The entire data communication is managed using ROS (Robot Operating System) which provides a robust platform for implementation of Robotics Projects. Jetson Nano is used to provide powerful on-board processing capabilities. Sensor fusion is performed on the data received from the different sensors to improve the accuracy of the decision making and inferences that we derive from the data. This data is then used to create a localized map of the environment. In this step, the position of the vehicle is obtained with respect to the Mapping done using the sensor data.
The different SLAM (Simultaneous Localization and Mapping) techniques used for this purpose are Hector Mapping and GMapping which are widely used mapping techniques in ROS. Apart from SLAM that primarily uses LIDAR data, Visual Odometry is implemented using a Monocular Camera. The sensor fused data is then used by Adaptive Monte Carlo Localization for car localization. Using the localized map developed, Path Planning techniques like "TEB planner" and "Dynamic Window Approach" are implemented for autonomous navigation of the vehicle. The last step in the Project is the implantation of Control which is the final decision making block in the pipeline that gives speed and steering data for the navigation that is compatible with Ackermann Kinematics. The implementation of such a control block under a ROS framework using the three sensors, viz, LIDAR, Camera and IMU is a novel approach that is undertaken in this project.

## 1. INTRODUCTION

The concept of automated driving systems has existed for over several decades now. Initially it was mainly used to control and maintain the speed of the vehicles. Later, it was gradually enhanced to make it autonomous in highway environments. Although there has been continuous research for several decades now, there has been no commercial "Self-driving Car" in the market currently. But all this is about to change significantly with the upcoming technologies like Modern Sensor Systems and Advanced Driver Assistance systems.

Self-Driving or Autonomous Driving can be implemented using multiple approaches, that is, by using pure classical control-system based approach, or by using purely a Deep Learning Methodology such as Behavioral Cloning. However, in the current commercially available self-driving cars, a combination of both these methods is used in conjunction with one another and is combined in a way that will complement each other. The possibility of damage caused could be catastrophic both to the users as well as everyone around. This necessitates the development to be carried out keeping all the stakeholders in mind. As the need for greater accuracy and foolproof design increases, there is a need to explore the problem in different ways.

## 2. HARDWARE

For the purpose of implementing Autonomous Driving, we have designed a scaled prototype car which has an Ackermann Steering System which is similar to most commercial cars.

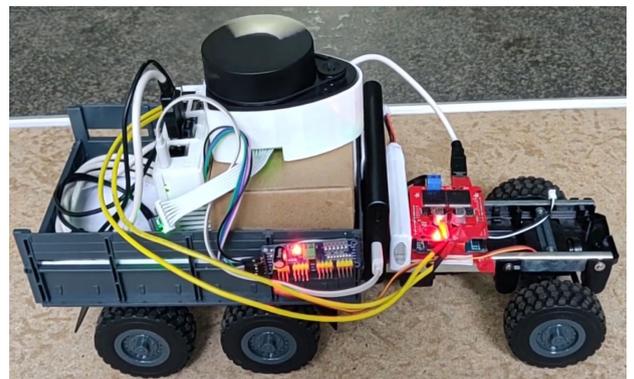

Fig1 :- Scaled Autonomous Prototype Car

The car is equipped with sensors like 2-D LIDAR, Camera, IMU and has Nvidia Jetson Nano for the on-board computing unit. Recent research in the field of Autonomous Vehicles has suggested LIDAR could be used effectively for obstacle detection and avoidance. [1] The car direction is controlled by a steering servo and the speed of the car is controlled by a Brushed DC motor. The car used is a powerful 6-wheel drive with all-wheel transmission line and has the necessary torque to be equipped with all the various hardware components

Table 1.Hardware Used.

|    | Main Hardware Components |
|----|--------------------------|
| 1. | RPlidar Sensor           |
| 2. | Nvidia Jetson Nano Computer |
| 3. | Raspberry Pi Camera      |
| 4. | IMU- MPU 9265            |

### 3. SELF-DRIVING PIPELINE

The Self-driving problem can be broken down into smaller abstract components for better understanding. There are four main components when it comes to Self-Driving/Autonomous Driving, i.e. *Perception, Localization, Path Planning* and *Control*. These four parts are implemented in succession and each step is dependent on the previous step. There this can be understood as the Self-Driving Pipeline.

#### 3.1. Perception

"Perception" is the First step in the pipeline which is responsible for collecting data of the outside environment in which the Car is supposed to navigate. This step involves using multiple sensors for better accuracy and redundancies. With this the self-driving car is able to "see" the environment.

#### 3.2. Localization

"Localization and Mapping" is the step where the Car will first accurately generate high precision maps of the environment and will attempt to 'localize' within that map, meaning it will determine where exactly it is situated in the map.

#### 3.3. Path-Planning

"Path Planning" is one of the most important steps in the entire pipeline where the Car will generate a "path" which contains a series of x-y coordinates along with orientation quaternions given a start point and a goal point. The Path Planner will keep in mind the kinematic constraints of the vehicle along with obstacle avoidance.

#### 3.4. Control

"Control" is the step where we implement a vehicle control system that will take in the path generated by the Path Planner and will attempt to control the car so that it follows that path as much as possible within a certain tolerance margin.

### 4. METHODOLOGY

This section gives an overview of the high-level methodology to achieve autonomous driving.

The first step is to generate odometry which is a requirement for various algorithms in further stages. Odometry is generated using all the different sensors in the car such as LiDAR, Camera and IMU.

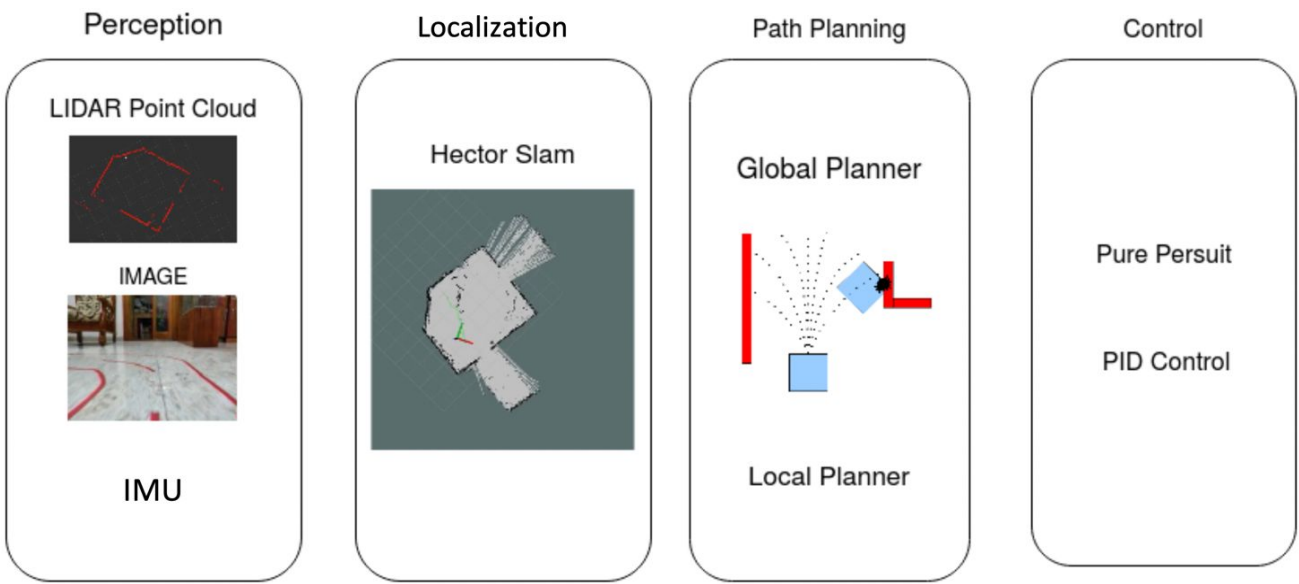

The next step is to generate a precise 2 dimensional Map of the environment the car will be navigating in. The map is generated using the technique of Simultaneous Localization and Mapping (SLAM). We have used and tested various SLAM algorithms such as Hector Mapping and G-mapping to generate the map of the environment. This step is very important since the map generated here will be used in subsequent steps.

After generating a precise map of the environment, the next task is to determine the position and orientation of the car with respect to this generated map. This is called '*Localization*''. Using different sensor streams and the map, the localizer will determine the pose of the car in the map. The localizer we have used is called the Adaptive Monte Carlo Localizer.

The next step in the methodology is Path Planning. This is the step in which actual autonomous navigation begins to take shape. In this step, the Path Planning algorithms will generate a trajectory from a Source point to the Destination Point(Goal Pose) lying on the map.

The final step is the Controller. The controller is responsible for generating driving parameters to the car, such as Steering angle and Velocity. The Path Planner supplies waypoints of the trajectory to the Controller, and the Controller will output suitable driving parameters in order to reach those waypoints.

## 5. IMPLEMENTATION

In this section, we give a detailed overview of the implementation of various algorithms used in the Pipeline.

### 5.1. Odometry

Odometry is the process of estimating the change in position over time with respect to an initial position using data from sensors placed in the car.

The usual way of obtaining odometry is through hall-effect based wheel/motor encoders. However we have not used that setup in our implementation for two reasons, the first reason is that encoded motors are expensive and even though adding external wheel encoders is possible, it requires specialized hardware

modifications.The second reason we have not used wheel encoders is because we have an Ackermann Kinematics based car. To use Wheel Encoders in an Ackermann Car, we require a very accurate simulated kinematic model to use wheel encoders for odometry.

The sources of odometry used in the are are:
1. R2FO Laser Scan Matching based Odometry
2. Visual Odometry
3. IMU dead-reckoning

### 5.1.1. Laser Scan Matching

RF2O is a fast and precise method to estimate the planar motion of a LIDAR from consecutive range scans. For every scanned point the range flow constraint equation is formulated in terms of the sensor velocity, and minimize a robust function of the resulting geometric constraints to obtain the motion estimate. In contrast to conventional approaches, this method does not search for correspondences but performs dense scan alignment based on the scan gradients, like a dense 3D visual odometry. The minimization problem is solved in a coarse-to-fine scheme to cope with large displacements, and a smooth filter based on the covariance of the estimate is employed to handle uncertainty in unconstrained scenarios (e.g. featureless environments).

### 5.1.2. Visual Odometry

Visual odometry is a technique where only cameras are used for generating Odometry Information. The main idea behind visual odometry is to estimate the motion of the camera in 3D space using time sequential images [10]. The Algorithm will compare series of time separated images to estimate the most probable motion undertaken in the camera frame. This is done by comparing the distance between the similar points in the consecutive images. [9] The visual odometry algorithm we have used is called ORB-SLAM-2.[11]

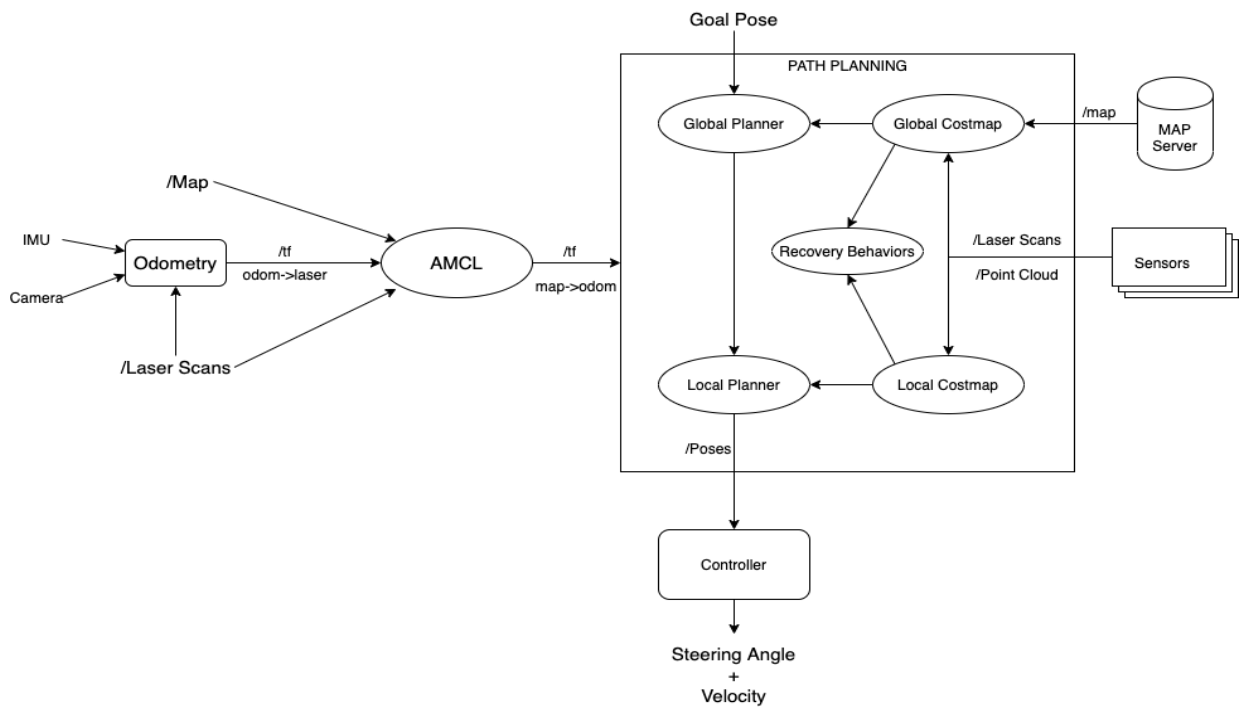

## 5.2. Simultaneous Localization and Mapping (SLAM)

For an autonomous vehicle to navigate in an environment, the perquisite is to have a map of that environment. To obtain this, we use a technique known as Simultaneous Localization and Mapping. SLAM is done to estimate the pose of a robot and the map of the environment at the same time.

The major techniques which use 2D LIDAR for SLAM are Gmapping, Hector SLAM and Cartographer. Based on the comparative studies [2] of these different methods, we chose to use Hector Mapping to implement SLAM. The Gmapping method does not provide reliable results. Whereas the Cartographer method was more suitable for more sophisticated LIDARs. The Hector Mapping algorithm was developed as open source modules that act as building blocks for systems performing Urban Search and Rescue (USAR) tasks. The details of the Algorithm and the modules used are explained in the [3].

This is implemented using *hector_slam* node [5] that consists of *hector_mapping, hector_map_server , hector_geotiff* and *hector_trajectory_server* modules. The system does not require odometry data, instead it relies on fast LIDAR data scan-matching at full LIDAR update rate. The mathematics behind the algorithm is explained in [4].

## 5.3. Particle Filter Localization

The Particle Filter Localizer is responsible for determining the position of the robot within the map created using Hector SLAM. The Particle Filter requires three inputs , that is , it needs a recorded map of the environment, Laser scans from the LIDAR and a source of Odometry. The combined odometry from the 3 sensors are given as odometry for the particle filter using the sensor fusion technique.

In general, a particle filter includes four steps :
  (a) Generating a set of particles
  (b) Measure the probability of each particle
  (c) Resample based on the probability weight [6]
  (d) Repeatedly move to approach orientation

A simple intuitive understanding of the Particle Filter localization is that given a current laser scan, the algorithm will try to match the current laser scan with the saved map and estimate the most probable location and orientation within that map. In our project we have used a variation of the Monte Carlo Localizer named as Adaptive Monte Carlo Localizer.

The algorithm is implemented on the ROS platform using the popular AMCL Library. [13] The five main nodes used in this algorithm are [7] :

a) *sample_motion_model_odometry*
b) *beam_ramge_finder_model*
c) *likelihood_feild_range_finder*
d) *augmented_MCL*
e) *KLD_sampling_MCL*

### 5.3.1. *sample_motion_model_odometry*

The above algorithm captures the relative motion information of the AV. The pose consists of the x-coordinate, y-coordinate and the angle of orientation. This is represented by the pose of the robot with respect to the robot's internal odometry.

Basically in this algorithm, the pose of the robot is found given the linear velocity $u_t$ and the previous pose of the robot, $x_{t-1}$. The $u_t$ by a three step process of finding the relative motion between the 2 consecutive poses. It involves a concatenation of three basic motions: a rotation, a straight-line motion (translation), and another rotation. It considers the inverse motion model to obtain this.

### 5.3.2. *beam_range_finder_model*

Rangefinders are among the most popular sensors in robotics. Rangefinders measure the range to nearby objects. Range may be measured along a beam—which is a good model of the workings of laser range finders.

This algorithm basically estimates the $p(z_t \mid x_t, m)$. That is it estimates the end point of a laser scan given the pose of the robot and the map. To achieve this the model takes into consideration mainly four types of errors to arrive at an accurate measurement algorithm. They are :

a) Correct range with local measurement noise

This considers the error on the account of the sensor to correctly measure the range. The error arises due to limited resolution of the sensors, atmospheric effect on the measurement signal and so on. In case of LIDARs this can be due to foggy environments.

b) Unexpected objects

Though the input map for the algorithm is static, the environment of the AV is dynamic. As a result objects that are not present in the map cause the detection of short ranges. With respect to AVs a common example are people that share the same environment.

To solve this problem it can either consider a separate state vector to dynamically keep track of each individual, which is both complex and computationally intensive, or consider their presence in the environment as a sensor noise that causes a shorter range. This makes the problem much simpler. Therefore the likelihood of sensing unexpected objects decreases with range, which gives us an excellent model for the problem.

c) Failures

Sometimes the obstacles are missed altogether. In case of LIDARs this can be due to bright outdoor sunlight and the presence of dark surfaces that can result in absorption of signal by the surface. This can be tackled by introducing an allowable level of $z_{max}$.

d) Random Measurements

The sensors may sometimes produce measurements that are completely unexplainable. This can be possibly caused due to phantom readings caused by bouncing off walls or due to cross talk between sensors.

### 5.3.3. likelihood_feild_range_finder

Unlike the other models that rely on conditional probability relative to a physical model of the sensor, this is an "ad-hoc" algorithm that lacks any such structure. Despite this it has been able to work well in practice. Here the end point measurements are done with respect to a global coordinate system that is obtained by performing a trigonometric transformation of the end point measurement. The errors considered for this model are similar to the ones mentioned in *beam_range_finder_model*.

### 5.3.4. Augmented_MCL

MCL in its basic form is incapable of recovering from robot kidnapping or global localization failure. This is due to the fact that all the particles are concentrated near the actual position of the robot, so when there is incorrect pose that is considered the robot completely loses control because there are no particles available to guide it back to the correct path.

To solve this problem random particles are added to the sample space to effectively recover the robot from such situations and over-ride the high concentration of the misguided robot. This is done by continuously comparing the short-term and the long-term average of measurement likelihood. What this does reveal is, if these two quantities are equal then the requirement of introducing the random articles is absent, that is, the robot is well in control. Whereas, a larger difference between these two quantities indicate a greater need to introduce random particles to enable course correction.

### 5.3.5. KLD_Sampling_MCL

Kullback-Leibler divergence sampling method is a variant of the MCL that statistically determines the number of particles required such that the error between the true posterior and the sample-based approximation is less than ε. This reduces the computational resources required to process the particles. It does this by dynamically adapting the number of the samples required over time. For this it takes into account the volume of the state space that is covered by the particles. This intuitive idea behind this is, there is no need for a large number of particles for the model when the information being provided by particles that are very closely placed are highly similar.

### 5.4. Path Planning

Path Planning is the process of generating feasible trajectories from a source point to a destination point taking into consideration various aspects such as dynamic obstacles, kinematics and geometry of the car. The path planner requires a '*costmap*' to compute the trajectories.

The costmap is an important step in the path planning process. The map obtained through SLAM is divided into smaller subcomponents by the use of a grid. The size and coarse of the grid can be specified and tuned in accordance to the physical parameters of the environment.

A costmap is a grid map where each cell is assigned a specific value or cost: higher costs indicate a smaller distance between the robot and an obstacle. To achieve this, we must specify the "base-footprint" which is nothing but the outer collidable boundary of the Autonomous Vehicle. Based on this each cell in the costmap can have one of 255 different cost values. The underlying structure that it uses is capable of representing only three. Specifically, each cell in this structure can be either free, occupied, or unknown. Another important concept in costmaps is Inflation.

Inflation is the process of propagating cost values out from occupied cells that decrease with distance. For this purpose, 5 specific symbols for costmap values are defined as they relate to a robot.

- "Lethal" cost:- real-world physical obstacle present in the cell. Probability of collision is 100%
- "Inscribed" cost:- Cells which are less than the car's inscribed radius away from a real-world physical obstacle. Probability of collision is high.
- "Possibly circumscribed" cost: - Cells which are less than the car's circumscribed radius away from an obstacle, also depends on the car orientation. Low Probability of collision.
- "Freespace" Cost:- Value of cost is zero, which means that region is devoid of obstacles and the robot can traverse in that region without concern
- "Unknown" cost is given to those cells for which we do not have any information about. The interpretation of that is left to the user.
- All the other cells are assigned with a cost that is in between "Freespace" and "Possibly circumscribed according to the distance from itself to a "Lethal" cell and the decay/gradient function provided by the user.

The high level path planning strategy used is called hierarchical planning. That is we split the entire problem of Path Planning into two components, that is , Global

Planner and Local Planner. By doing this we can simplify the overall problem and also reduce the computation required.

### 5.4.1. Global Planner

Global Planner deals with generating the high-level trajectory/path from the source to destination. The Global Planner uses the costmap to determine the most optimized path with the least cost from source to destination. The Global Planner only takes into consideration the costmap and the footprint of the car, it does not consider the kinematics or the geometry of the car while generating the path. This is the reason a local planner is required.

In our implementation we have used Dijkstra's algorithm as the global path setter. Dikstra's algorithm was was chosen over others because in terms of speed and space efficiency, and simplicity, Dijkstra's algorithm has outperformed the other path planning algorithms such as Breadth First and A*.

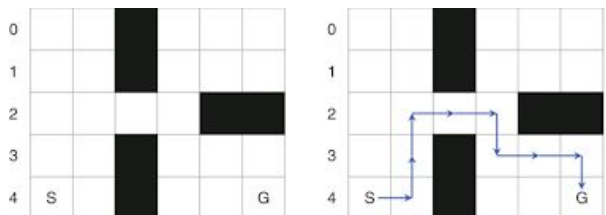

Fig. 4:- Dijstra's path example in a costmap

### 5.4.2. Local Planner

As explained previously, the approach to path-planning here is hierarchical. The top level planning is performed by the Global Planner and the low level planning, taking into consideration various other constraints is performed by the Local Planner. The Local Planner uses the Global path generated by the Dijstra's algorithm to come up with the Local Path.

The Global Path generated is converted into a set of waypoints which is then given as input to the local planner. The algorithm that we have used for the Local Planner is the Time-Elastic-Band Local Planner[8].

The normal "elastic band" transforms a path generated by a global planner avoiding contact with obstacles and with respect to the shortest path length. It does not take into consideration dynamic constraints of the underlying system.

'Time elastic band' optimizes the car's trajectories by making suitable modifications to the trajectory generated initially by the global planner. Some of the important parameters considered while performing this trajectory optimization are the overall path length, trajectory execution time, separation from obstacles, intermediate waypoints and compliance with the car's dynamic, kinematic and geometric constraints. The TEB planner especially considers the spatial-temporal dynamic constraints such as limited car velocities, turning radius and acceleration. Like stated previously, the main purpose for TEB Local Planner is to generate kinematically compatible trajectories for the car, taking into account obstacles and other optimization parameters.

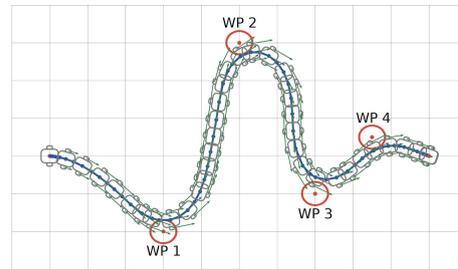

Fig.54:- TEB Trajectory for given waypoints

## 6. SENSOR FUSION

Sensor fusion is the process of bringing together the inputs from multiple sensors like lidars, cameras and IMU's in order to form a single model or image of the environment around a vehicle. It will balance the strengths of various sensors and hence, the resulting model will be more accurate. The autonomous vehicle will utilize the information provided through sensor fusion to perform more-intelligent actions.

In our implementation we have used the Extended Kalman Filter to fuse different sources of odometry into a single odometry source which is more precise. The different sources of odometry that are fused are :-

1. R2FO Laser Scan Matching based Odometry
2. Visual Odometry
3. IMU dead-reckoning

Also the output pose from the AMCL is finally fused with the pose estimate from the odometry to give the final car pose in the map.

## 6. CONTROLLER

The controller is the final stage in the self-driving pipeline. The function of the controller is to generate driving parameters given the path from the local planner. The controller is responsible for making the car achieve the series of poses given by the TEB planner. A pose consists of an x,y coordinate and an orientation. The controller used is dependent on the kinematics of the car. In our implementation, we have used the Ackermann steering system. This is because we want to emulate the real-world commercial cars in use today, which also use The Ackermann Steering system.

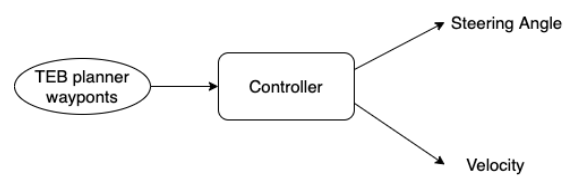

Fig.6:- Controller Block Diagram

The Controller is implemented using a simple PID control system that will output linear velocity and angular velocity required to achieve each consecutive waypoint.

However the controller we have used is defined for a two wheel differential drive robot that can turn in place. Therefore we must perform an additional stage of operation to convert the commands for an Ackermann Steering based car that accepts a steering angle instead of angular velocity.

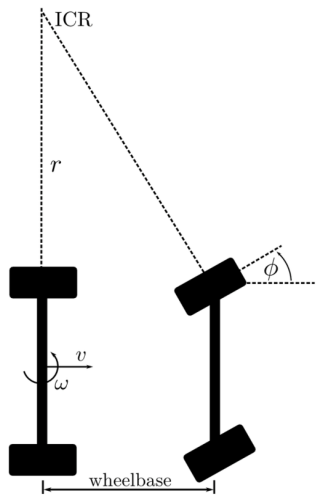

Fig.7 :- Ackerman Geometry

The relevant variables including the translational velocity $v$ and the steering angle $\phi$ are illustrated in the following figure. ICR denotes the instant centre of rotation. For a vanishing angular velocity $\omega=0$ the turning radius r reaches to infinity which will lead to a zero steering angle $\phi=0$. Alternatively, the turning radius r might be computed by $v/\omega$. Then, the steering angle can be obtained by

$$\phi=\operatorname{atan}(\text{wheelbase}/r).$$ (will change in word)

Therefore, by using the above transformation, we have computed the final steering angle that is sent to the servo, and the linear velocity component does not require any transformation and is directly fed to the Motor in the form of a PWM value.

## 7. CONCLUSION

In this paper, we describe the end-to-end working and implementation of a Self-Driving Car. We have successfully demonstrated a proof of concept self-driving car by implementing existing methods and sensors into the project like LIDAR used in modern self-driving cars in the form of a scaled prototype.